\documentclass{article}

\usepackage[english]{babel}

\usepackage[letterpaper,top=2cm,bottom=2cm,left=3cm,right=3cm,marginparwidth=1.75cm]{geometry}

\usepackage{amsmath}
\usepackage{graphicx}
\usepackage[colorlinks=true, allcolors=blue]{hyperref}

\usepackage{caption}
\usepackage{subcaption}
\usepackage{ulem}

\title{Truck Axle Detection with Convolutional Neural Networks}
\author{Leandro Arab Marcomini, Andre Luiz Cunha}

\begin{document}
\maketitle

\begin{abstract}

Axle count in trucks is important to the classification of vehicles and to the operation of road systems. It is used in the determination of service fees and in the impact on the pavement. Although axle count can be achieved with traditional methods, such as manual labor, it is increasingly possible to count axles using deep learning and computer vision methods. This paper aims to compare three deep-learning object detection algorithms, YOLO, Faster R-CNN, and SSD, for the detection of truck axles. A dataset was built to provide training and testing examples for the neural networks. The training was done on different base models, to increase training time efficiency and to compare results. We evaluated results based on five metrics: precision, recall, mAP, F1-score, and FPS count. Results indicate that YOLO and SSD have similar accuracy and performance, with more than 96\% mAP for both models. Datasets and codes are publicly available for download.

\end{abstract}

\section{Introduction}

In transportation systems, the number of axles of a heavy vehicle is essential to determine limitations on circulation, to avoid damages to the roads, or to compose service fees, as compensation for the long-term damage a vehicle can incur on the pavement \cite{hatmoko2019investigating}. That is why it is important to know exactly the number of axles a commercial vehicle, such as a truck, has. A fully automated system, capable of identifying axles, is important because it allows for 24-hour monitoring and data collection, without the need for human resources to be invested. Monitoring truck axles also helps with law compliance, making roads safer for all drivers.

In Brazil, truck axle counts are manually acquired, with an operator inputting information. On toll plazas, where axle numbers are used to compose a service fee for heavy vehicles, the operator has to count the axles while the vehicle is approaching, with a limited view angle, creating space for human error \cite{lee1990infrared}. Operators are also constantly exposed to exhaust fumes, posing a threat to their health \cite{choi2016estimates}, especially in a country where 60\% of freight cargo is transported by trucks \cite{araujo2021freight}. 

Operators also manually input the number of axles in truck scales, used for law compliance. Since Brazil has strict rules for maximum weight on combinations of axles, as it is possible to see in Table~\ref{table:limites}, acquiring this information reliably is vital to guarantee law is followed, and no further damages are made to the pavement or in detriment to the overall road safety. In both examples, whether being on the calculation of fees or on imposing a non-compliance fine for drivers, the quality of inputted data is the key in the determination of the output. Manual input also contributes to a slower overall operation, decreasing the driver's perceived quality of service.

\begin{table}[h]
	\centering
	\caption{Schematics of the side view and the bottom view of a truck, with axle combinations and weight limits, in tonnes, in Brazil \cite{lei:denatran505}. Gray axles denote a single-wheel combination. }
	\label{table:limites}
	\begin{tabular}{|c|c|c|} 
	    \hline
		\textbf{Truck Axle - Side View} & \textbf{Truck Axle - Bottom View} & \textbf{Weight Limit (t)}\\ 
		\hline
		\raisebox{-\totalheight}{\includegraphics[width=.12\textwidth]{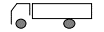}}    & \raisebox{-\totalheight}{\includegraphics[width=.12\textwidth]{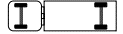}}  & \raisebox{-\totalheight}{6 + 6 = 12} \\
		\hline
		\raisebox{-\totalheight}{\includegraphics[width=.12\textwidth]{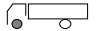}}    & \raisebox{-\totalheight}{\includegraphics[width=.13\textwidth]{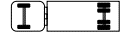}}  & \raisebox{-\totalheight}{6 + 10 = 16} \\
		\hline
		\raisebox{-\totalheight}{\includegraphics[width=.12\textwidth]{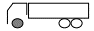}}    & \raisebox{-\totalheight}{\includegraphics[width=.12\textwidth]{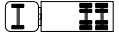}}  & \raisebox{-\totalheight}{6 + 17 = 23} \\
		\hline
		\raisebox{-\totalheight}{\includegraphics[width=.15\textwidth]{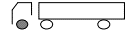}}    & \raisebox{-\totalheight}{\includegraphics[width=.15\textwidth]{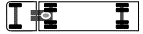}}  & \raisebox{-\totalheight}{6 + 10 + 10 = 26} \\
		\hline
		\raisebox{-\totalheight}{\includegraphics[width=.15\textwidth]{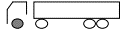}}    & \raisebox{-\totalheight}{\includegraphics[width=.15\textwidth]{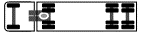}}  & \raisebox{-\totalheight}{6 + 10 + 17 = 33} \\
		\hline
		\raisebox{-\totalheight}{\includegraphics[width=.15\textwidth]{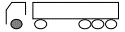}}    & \raisebox{-\totalheight}{\includegraphics[width=.15\textwidth]{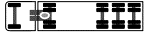}}  & \raisebox{-\totalheight}{6 + 10 + 25.5 = 41.5} \\
		\hline
		\raisebox{-\totalheight}{\includegraphics[width=.15\textwidth]{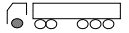}}    & \raisebox{-\totalheight}{\includegraphics[width=.15\textwidth]{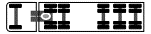}}  & \raisebox{-\totalheight}{6 + 17 + 25.5 = 48.5} \\
		\hline
	\end{tabular}
\end{table}

As the availability of vast amounts of data on the web increases, coupled with better performance of GPUs, object recognition with convolutional neural networks (CNN) \cite{al2017review} are increasingly applicable
in different areas of engineering. From Civil Engineering, in detecting damages due to vibration \cite{avci2022overview}, to Mechanical Engineering, observing engine cycles in search of abnormalities \cite{shahid2022real}, CNNs are helping to solve vision-based problems. In Transportation Engineering, CNNs are being used to detect potholes \cite{dewangan2021potnet}, to predict passenger flow in buses \cite{liu2020automatic}, and to detect and track vehicles \cite{maity2021faster}, among many other applications.

Because truck axle detection is a vision-based problem, it is possible to automate the task of extracting counts using a previously trained CNN. Using such a system decreases the time in which each truck is processed in a weighting system or in a toll plaza. It also contributes to the cost of cargo freight, increasing the overall fuel efficiency of transport.

In this paper, three popular object detection models based on deep learning techniques were analyzed and compared, namely YOLOv3, SSD, and Faster R-CNN, evaluating the performance through five metrics: (1) precision (2) recall (3) mAP, that takes into consideration the precision of the bounding box in predicting the location of the object, (4) F1-score, a balanced metric between precision and recall, (5) processing speed, to measure performance on modern hardware. Since truck axles are not the focus in commonly found image datasets on the web, it was necessary to create a set of images that would suffice to train and test all models. This labeled dataset was made available in a public repository, and it is one of the major contributions of this paper.

\section{Vision-based Axle Detection}

Traditional vision-based axle detection systems use a combination of border detectors with spatial analysis, to determine the position of axles and to exclude false positives. In \cite{grigoryev2015vision}, the authors adapt a Viola-Jones Algorithm for face detection and apply it to vehicle wheels, with spatial analysis and Hough Transform to remove any false detections. In \cite{panice2018metodo}, the Hough Transform algorithm is used to detect circles in truck images, with spatial analysis to determine the most likely position of axles. Recently, in Li et al. \cite{li2021research}, the authors applied HOG and SVM to detect gradients in the images to locate vehicle wheels in weight-in-motion systems.

The downside of traditional methods to extract and to detect features is that they do not take into consideration any form of image context. So, if a round object appears in an unlikely position, border detectors will identify them. Traditional solutions use a spatial system to aid in limiting the area of detection, or to exclude false detections, increasing processing times and cost. The use of algorithms that identify the objects themselves, and not just features, such as deep learning and neural networks, are trying to fix that. In \cite{gothankar2019circular}, the authors used a combination of Hough Transform and CNNs to increase precision in wheel detections. In \cite{li2021video}, the authors used the YOLOv5S to count axles and to classify vehicles in several categories. 

Since there are various deep learning algorithms, each with its advantages and disadvantages, this paper has the objective to compare popular Deep Learning methods, specifically CNNs, using performance metrics, to analyze and evaluate different algorithms to detect truck axles. In addition, this project should provide significant benefits to all areas because it contributes with a labeled dataset of truck axles, since nothing similar was found in the image databases available on the internet.

\section{Deep Learning Algorithms}

Research on object recognition systems with the use of deep learning techniques is leading to better overall accuracy and precision performances. Region-based detectors, such as Faster R-CNN \cite{ren2015faster}, which are notorious for low processing speeds, are increasingly reaching higher performances while maintaining good results - 36 FPS with 93\% mAP on vehicle detection \cite{kim2020comparison}. Single-stage detectors, such as YOLO \cite{redmon2016you}, and SSD \cite{liu2016ssd}, which on the other hand, have better performances in detriment of precision and recall, are improving their detection results (98\% mAP and 82 FPS for YOLO, 90\% mAP and 105 FPS for SSD, both on vehicle detection \cite{kim2020comparison}) with fast processing speeds, surpassing the needs of real-time applications.

\subsection{Faster R-CNN}

Convolutional Neural Networks (CNNs) use mathematical operations known as convolutions to process the pixels of an image in order to highlight specific characteristics used in object detection. Depending on the convolutional operation chosen, a neural network layer can highlight borders, and increase the difference between pixel value changes on an image, among many other filter options \cite{Goodfellow-et-al-2016}. 

\begin{figure}[h]
\centering
\includegraphics[width=\textwidth]{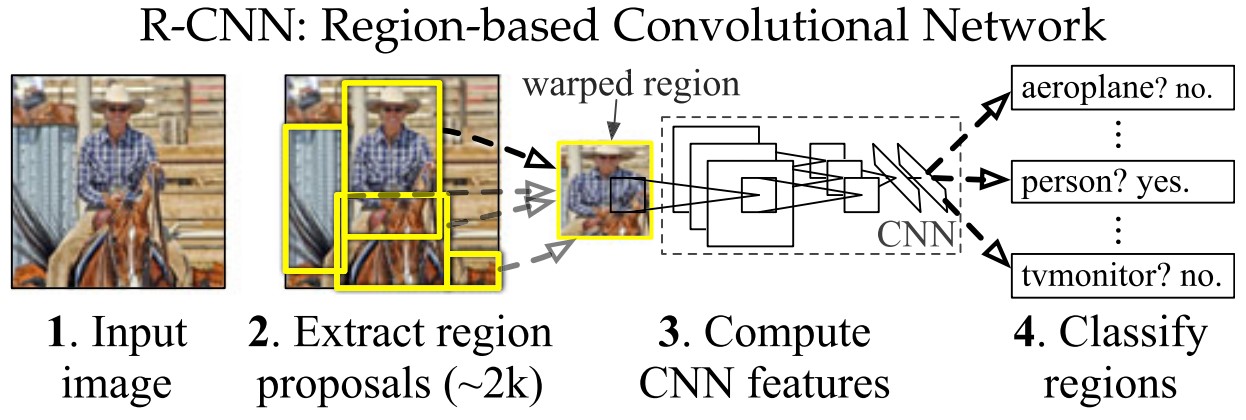}
\caption{Three main modules of an R-CNN network, with the proposal of approximately 2000 regions, followed by feature extraction, and classification on any of the available trained classes. Source: adapted from \cite{girshick2015region}.}
\label{fig:rcnnarq}
\end{figure}

Region-based Convolutional Neural Networks (R-CNN) \cite{girshick2015region} solve an important bottleneck with CNNs, which is the proposal of detection regions for objects with variable sizes, by using selective search. On Figure~\ref{fig:rcnnarq}, it is possible to notice the architecture of an R-CNN network, with variable-sized extracted region proposals. The features are computed only inside proposed regions, saving processing time and improving performance.

\begin{figure}[h]
\centering
\includegraphics[width=\textwidth]{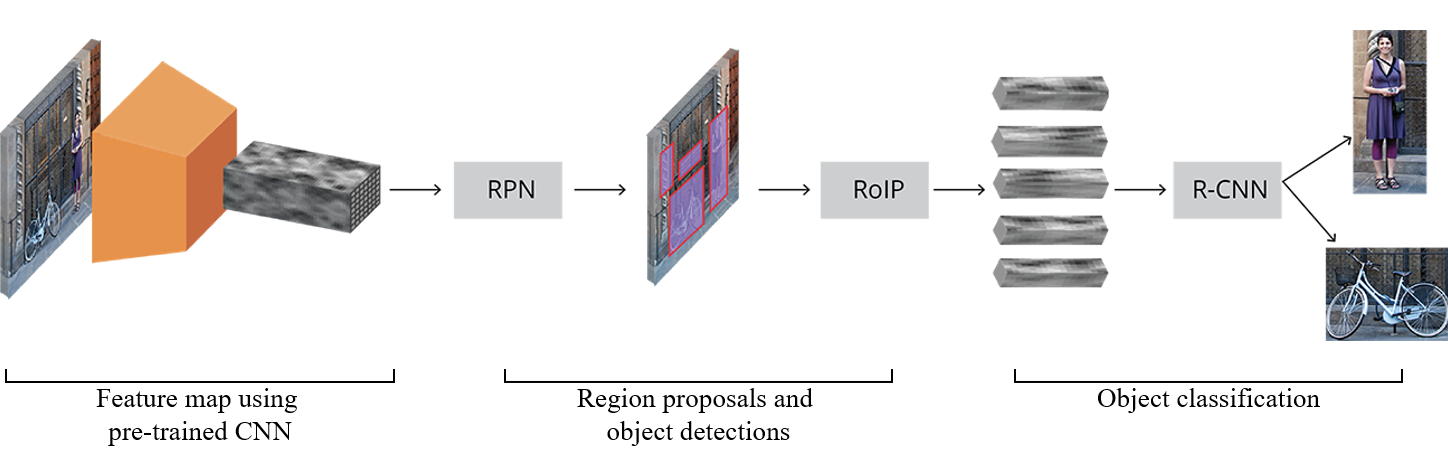}
\caption{Steps in detecting and classifying objects with Faster R-CNN. A pre-trained CNN extracts a feature map, which the RPN uses to propose detection regions and bounding boxes. The next CNN then classifies the objects. Source: adapted from \cite{frcnn-rpn-2018}.}
\label{fig:frcnn-rpn}
\end{figure}

On improved versions of R-CNN models, Fast R-CNN \cite{girshick2015fast} and Faster R-CNN \cite{ren2015faster}, selective search is replaced by a neural network specialized in region proposal - Region Proposal Network (RPN). The RPN uses a feature map extracted from an intermediate layer of a pre-trained CNN to determine anchor points. From these anchor points, the RPN fixes variable-sized bounding boxes to try to encompass objects. With that information, the subsequent CNN can decide whether objects are inside the regions, and fine-tune the location of the bounding boxes. It also classifies detected objects. The network architecture, with the use of RPN, can be seen in Figure~\ref{fig:frcnn-rpn}. This results in better performance, going from 50 seconds per image classification on the original paper to 0.2 seconds per image \cite{kim2020comparison}.

\subsection{YOLO}

\begin{figure}[h]
\centering
\includegraphics[width=\textwidth]{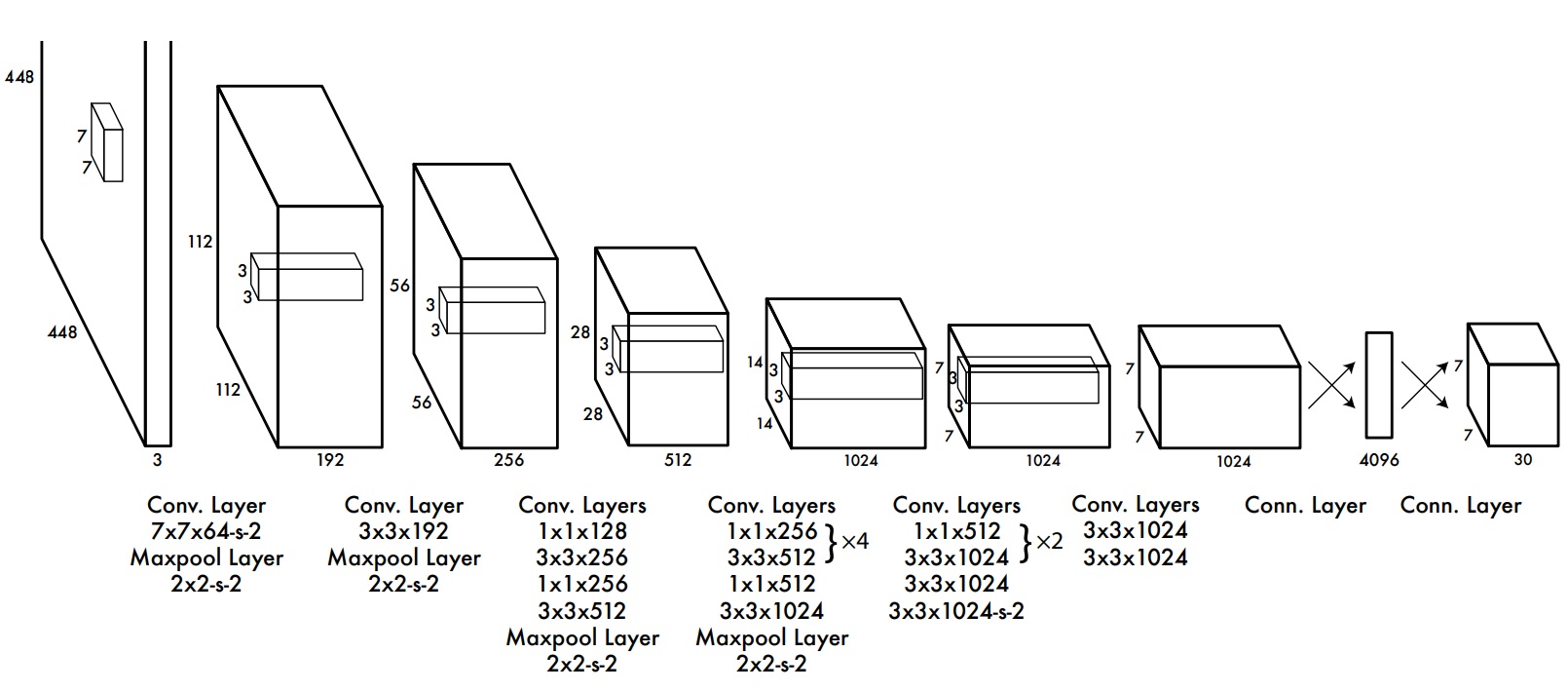}
\caption{YOLO architecture, with all convolutional layers and their sizes, and max pooling between layers to decrease data size. Source: adapted from \cite{redmon2016you}.}
\label{fig:yolo-original}
\end{figure}

You Only Look Once (YOLO) is a CNN with a single-stage detector. It is able to classify and infer bounding boxes of objects with a single pass through the pixels of an image. It uses ReLU (Rectified Linear Unit) as an activation function for inner neural network layers, with values below zero being passed as zero, and values greater than zero being passed without changes for the next layer. To increase it's processing speeds YOLO uses max pooling, lowering the amount of data between each layer \cite{redmon2016you}. The complete architecture of the original YOLO version can be seen in Figure~\ref{fig:yolo-original}.

\begin{figure}[h]
\centering
\includegraphics[width=\textwidth]{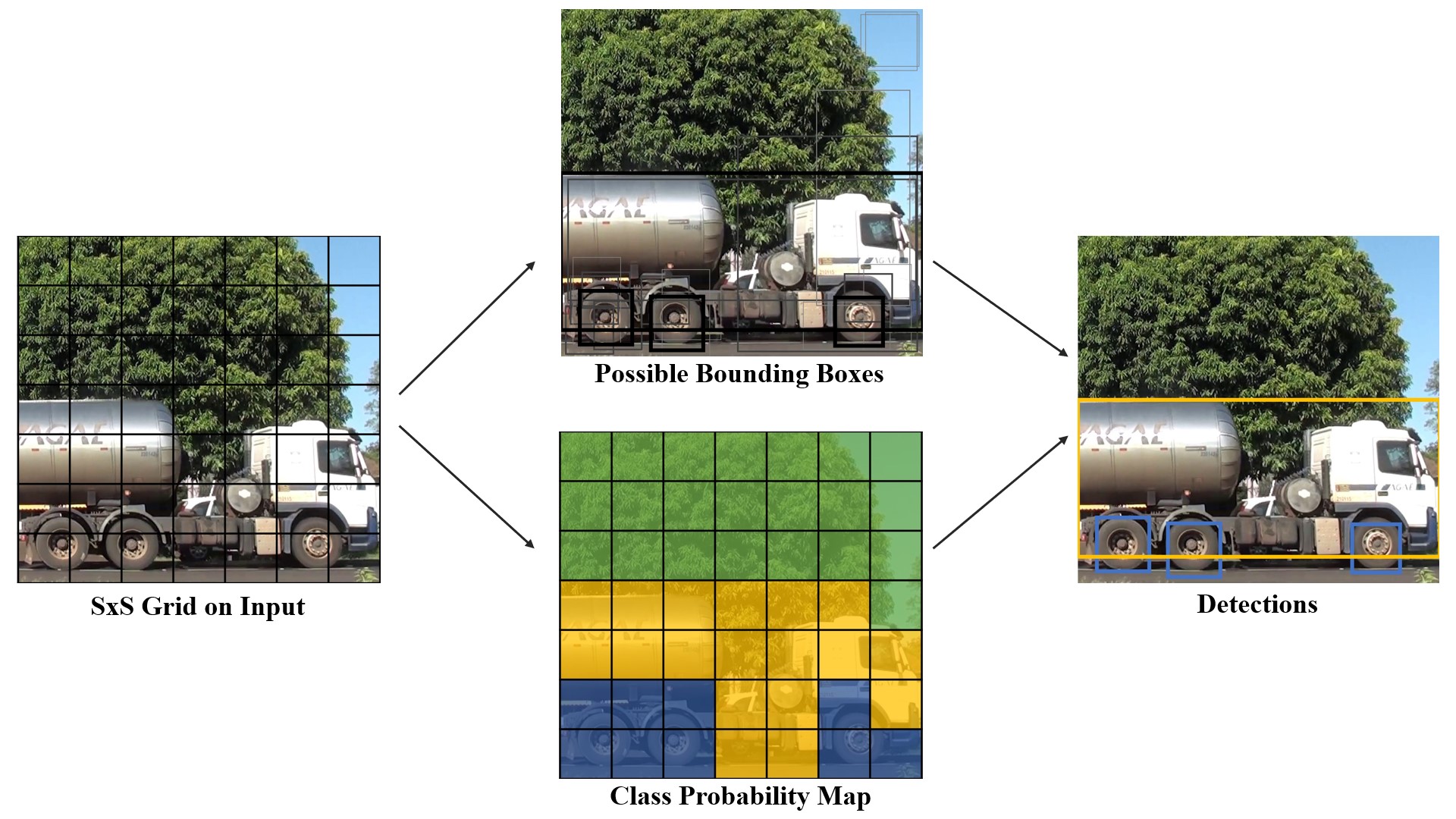}
\caption{YOLO Detection Sequence, with SxS grid on the input, class probability map, and bounding box determination.}
\label{fig:yolo-seq}
\end{figure}

To detect an object, YOLO divides the input image into an SxS grid. Each grid cell is responsible to predict bounding boxes and probabilities of any objects with their centroid in it. It is also responsible to calculate the class probability of that object, based on the training classes. That way, YOLO can detect all objects in an image in a single detection stage. It is possible to see the detection sequence in Figure~\ref{fig:yolo-seq}.

Since the release of the first YOLO version, many updates followed \cite{yolov2, yolov3, yolov4, yolov5}, with improved detection results and mAP scoring. Using the VOC2007 dataset as a training and testing base, YOLO scored 63.4\% mAP \cite{redmon2016you}, while YOLOv3 scored 74.5\% and YOLOv5, 82.7\% \cite{zhu2021improving}, a 30\% total increase in mAP performance. Recently, the introduction of YOLOv7 \cite{yolov7_2022} and YOLOv8 \cite{yolov8_2023} further increased accuracy and detection speed, with 30\% better results overall \cite{yolov8comparison}.

\subsection{SSD}

Similarly to YOLO, Single Shot Detector (SSD) \cite{liu2016ssd} applies a grid to the image, making each cell responsible to locate and classify objects. Without the use of sliding windows to locate objects, which is computationally intensive, processing times are reduced. Cells that have no objects are considered background and are no longer processed, further increasing performance, while cells with objects continue down the network.  

\begin{figure}[h]
\centering
\includegraphics[width=\textwidth]{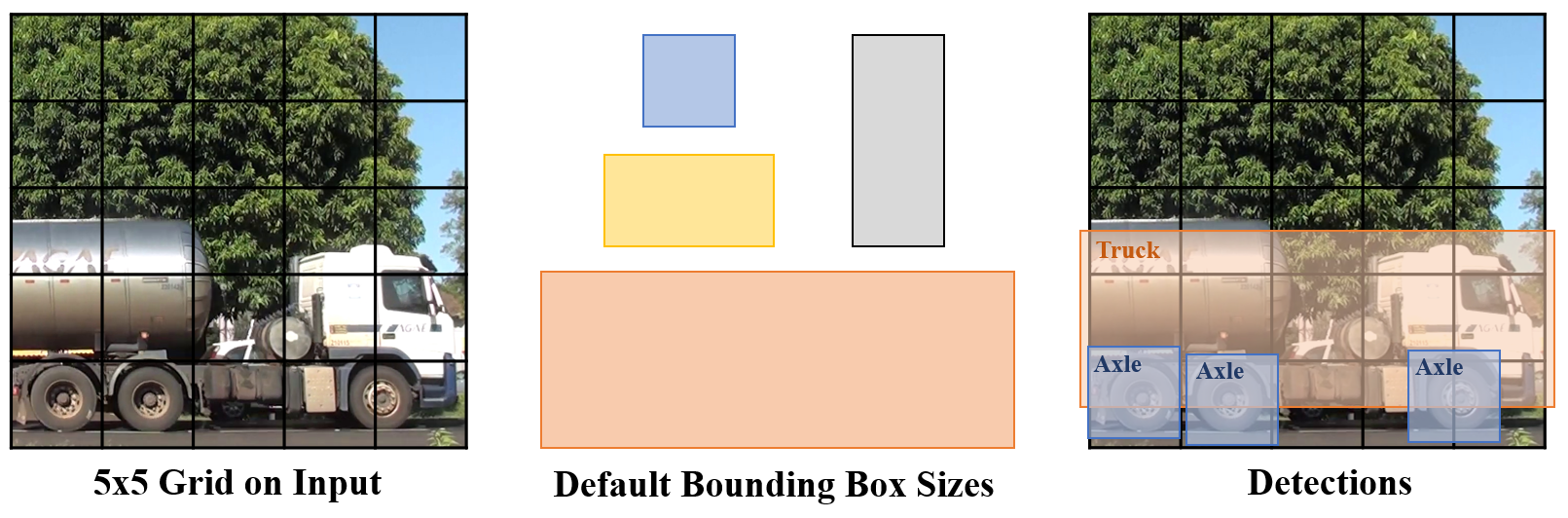}
\caption{An image is divided into a grid. Empty cells, with no detected objects, are classified as background. Cells with objects match a bounding box from a default set of options, classifying the object and locating it on the image.}
\label{fig:ssd-grid}
\end{figure}

\begin{figure}[h]
\centering
\includegraphics[width=\textwidth]{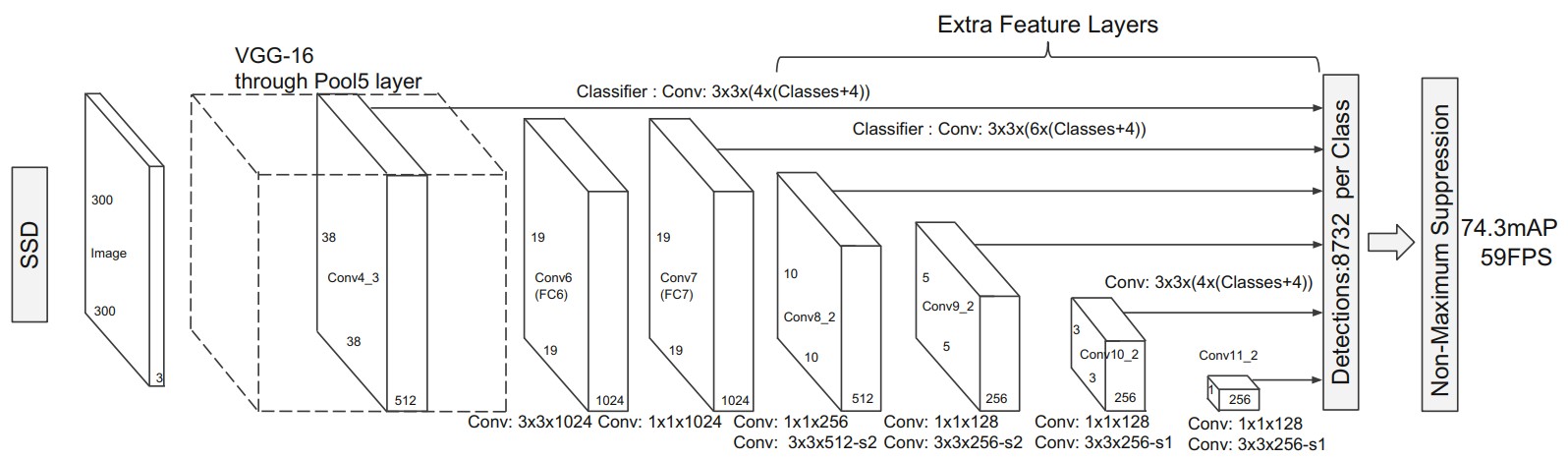}
\caption{SSD network architecture, with sequenced smaller convolutional layer sizes. Source: adapted from \cite{liu2016ssd}.}
\label{fig:ssdarq}
\end{figure}

 Detection and classification of objects are improved by removing the necessity of calculating correct bounding box sizes, limiting it to a choice between a set of default options. Bounding boxes are matched with objects in the training phase, taking into consideration not only the size but at what zoom level the object fits in that box during training. Figure~\ref{fig:ssd-grid} shows a simplified diagram of how the SSD model applies the grid to an image and fits the bounding boxes to the objects. Because the network uses sequentially smaller layers, objects with different sizes are detected \cite{liu2016ssd}. SSD is also another example of single-stage detection, processing the image only once. Figure~\ref{fig:ssdarq} shows the architecture of the SSD network, with varying convolutional layer sizes, to detect different-sized objects.   

SSD has also improved since its release, with the inclusion of context-based knowledge \cite{fu2017dssd, cao2018feature} to increase detection accuracy, going from 77.2\% mAP to 78.9\% on the PASCAL VOC2007 dataset, with focus on objects with different sizes.

\section{Experiment}

For the comparison of all methods, the experiment was divided in four parts (1) building a dataset (2) training the Neural Networks with a fixed training set of images (3) testing the resulting models in a testing set of images (4) evaluation using five different metrics.

\subsection{Dataset}

For this paper, a dataset with unobstructed images of truck axles was necessary. While sets of images with the purpose of training neural networks are widely available on the internet, a thorough search did not obtain effective results. Therefore, an extra objective was set: to share the dataset created for future research in this field.

\begin{figure*}[h]
	\begin{minipage}[t]{\textwidth}		
		\centering
		\includegraphics[width=0.985\textwidth]{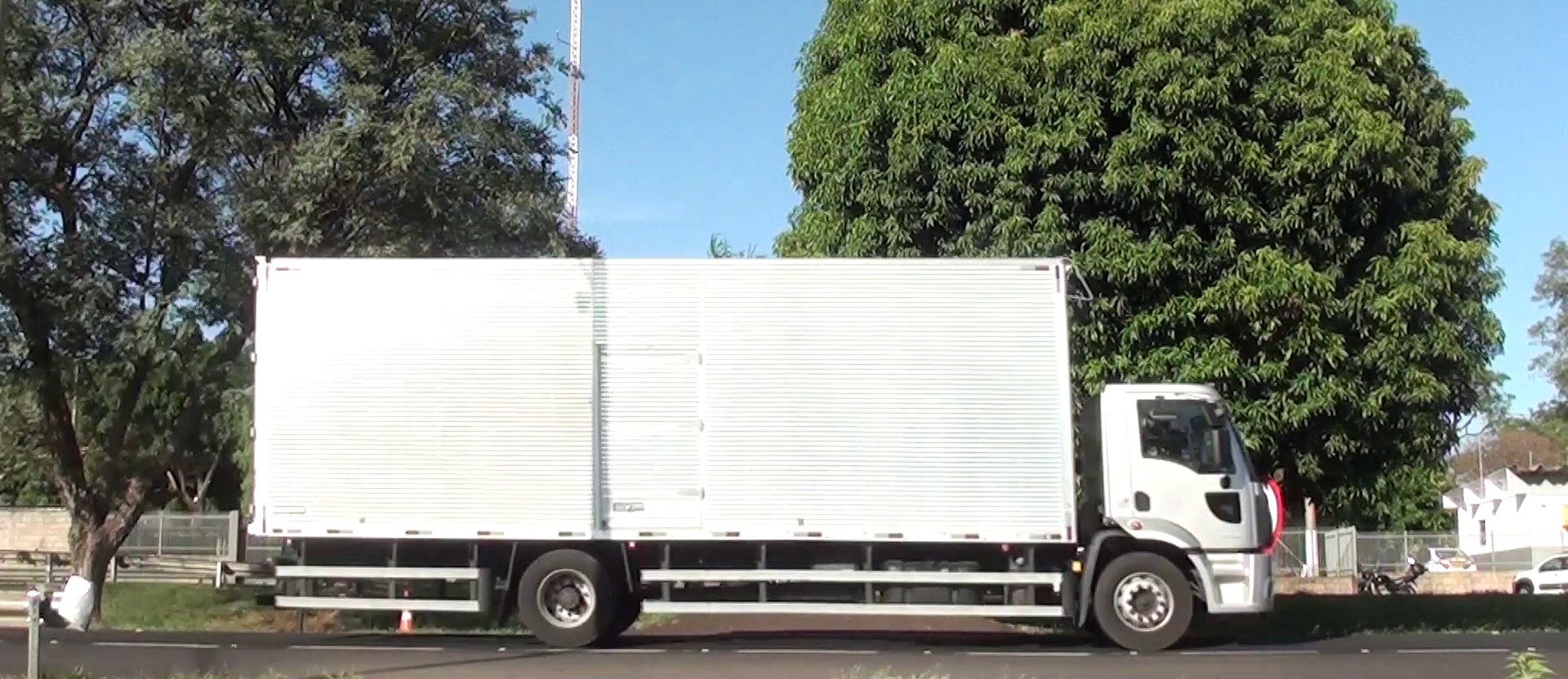}
		\label{fig:sub:subfigure2a}
	\end{minipage}
	\begin{minipage}[t]{\textwidth}
		\centering
		\includegraphics[width=0.985\textwidth]{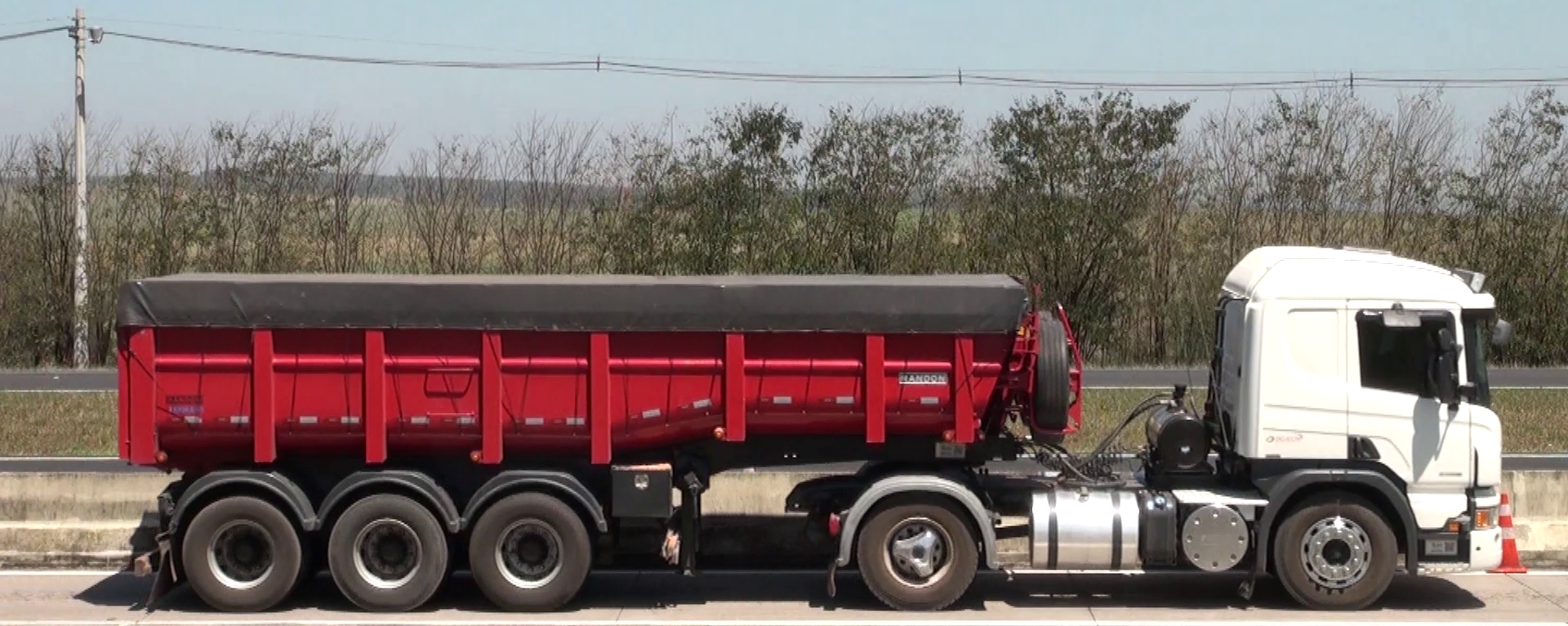}
		\label{fig:sub:subfigure2b}
	\end{minipage}
	\caption{Example of trucks recorded perpendicular to the road lane, to allow for axle capture.}
	\label{fig:trucks}
\end{figure*}

The dataset created in this paper contains 384 images of trucks manually extracted from videos recorded on highways in Brazil. All videos were recorded with cameras placed perpendicular to the lanes, on the right-side lateral clearance, as shown in Figure~\ref{fig:trucks}.

Since the object of interest is the axle, the sideways view from the truck was chosen to minimize distortion and to allow for the complete capture of the whole set of axles in a single truck. The distance from the road varies from video to video, but the goal of the placement is always to allow for the largest trucks to be entirely contained in a single frame of the video. The dataset is available for download on \cite{leandro_arab_marcomini_2021_5744737}, although temporarily without annotations.

\subsection{Training}

Axles are not common objects to appear as detection targets in pre-trained models available on the web. Because of that, it is necessary to train the neural network models and to supply as many axle images as possible.

All Neural Networks evaluated were trained with the same set of images from the dataset. The dataset was split so that 90\% of images were used to train the models, resulting in 346 images. The remaining 36 images were used for testing. All 346 images were manually annotated, resulting in 1184 truck axles. The annotation tool used was LabelMe \cite{labelme}. The dataset characteristics can be seen on Table~\ref{table:tab1}.

\begin{table}[h]
	\centering
	\caption{Training and testing datasets, with the number of images and axles.}
	\label{table:tab1}
	\begin{tabular}{|c|c|c|} 
	    \hline
		& \textbf{Training Set} & \textbf{Testing Set}\\ 
		\hline
		\textbf{Images} & 346  & 36\\
		\hline
		\textbf{Axles}  & 1184 & 119\\
		\hline
	\end{tabular}
\end{table}

In order to speed up the training process, it is possible to use pre-trained weight files and iterate over them as a base for the model. For the YOLO network, three different pre-trained weight files were used: YOLOv3-tiny, YOLOv3-416, and YOLOv3-spp. These specific base models were chosen because of the differences in performance results in \cite{yolov3} and \cite{yolov3site}. For both the Faster R-CNN network and the SSD network, only one weight file was used as a base for training, the ResNet50 \cite{he2016deep} and the SSD300 \cite{ssd2016}, respectively.

\subsection{Evaluation}

\begin{figure}[h]
\centering
\includegraphics[width=\textwidth]{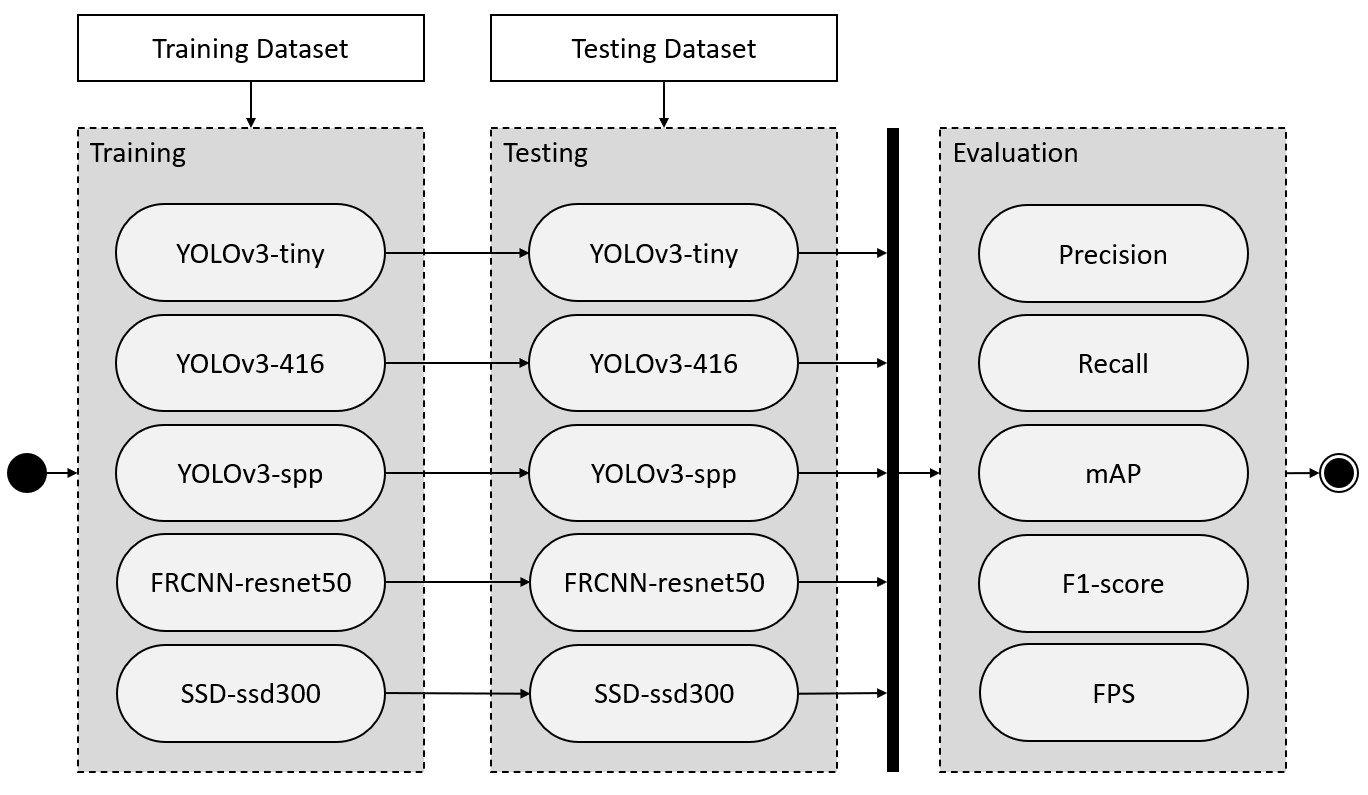}
\caption{Flowchart for the applied method, with training and testing the neural networks, and the evaluation of results based on five different metrics.}
\label{fig:fluxo}
\end{figure}

The resulting five different neural network models were evaluated by comparing accuracy and precision, to measure the performance related to finding and classifying the objects, with the application of mAP and F1-score metrics. The processing speed was also measured, with the amount of frames per second each model was capable of processing. A flowchart with the applied method and steps can be seen on Figure~\ref{fig:fluxo}.

\subsection{Experimental Results}

\begin{figure}[h]
    \centering
	\begin{subfigure}[b]{0.985\textwidth}		
		\centering
		\includegraphics[width=\textwidth]{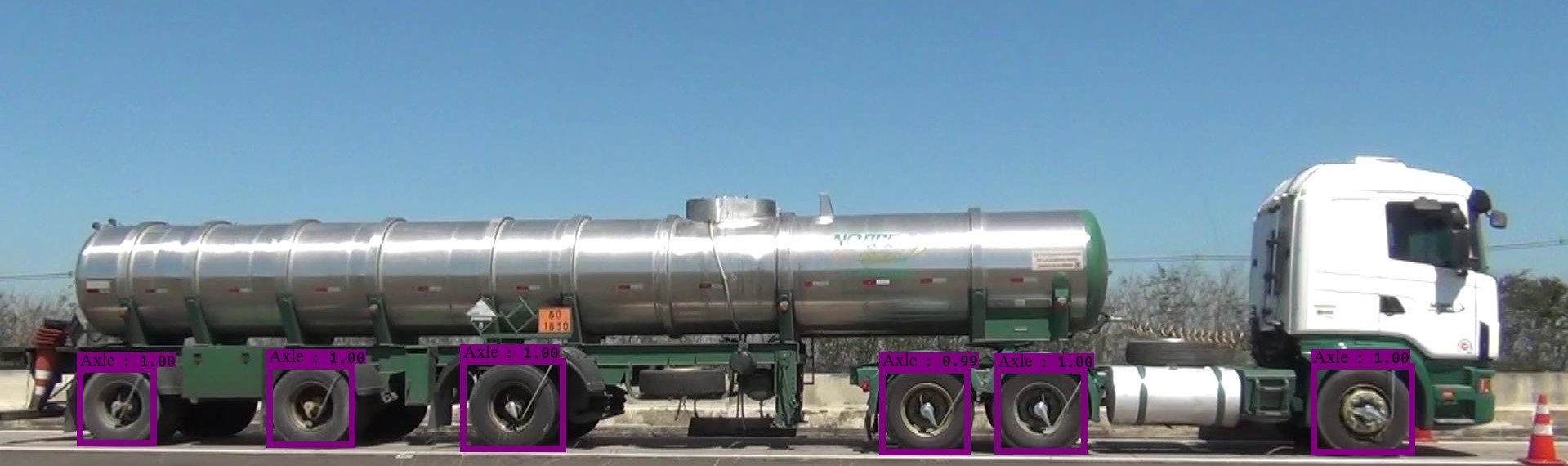}
		\caption{}
		\label{fig:sub:subfiga}
	\end{subfigure}
	\begin{subfigure}[b]{0.985\textwidth}		
		\centering
		\includegraphics[width=\textwidth]{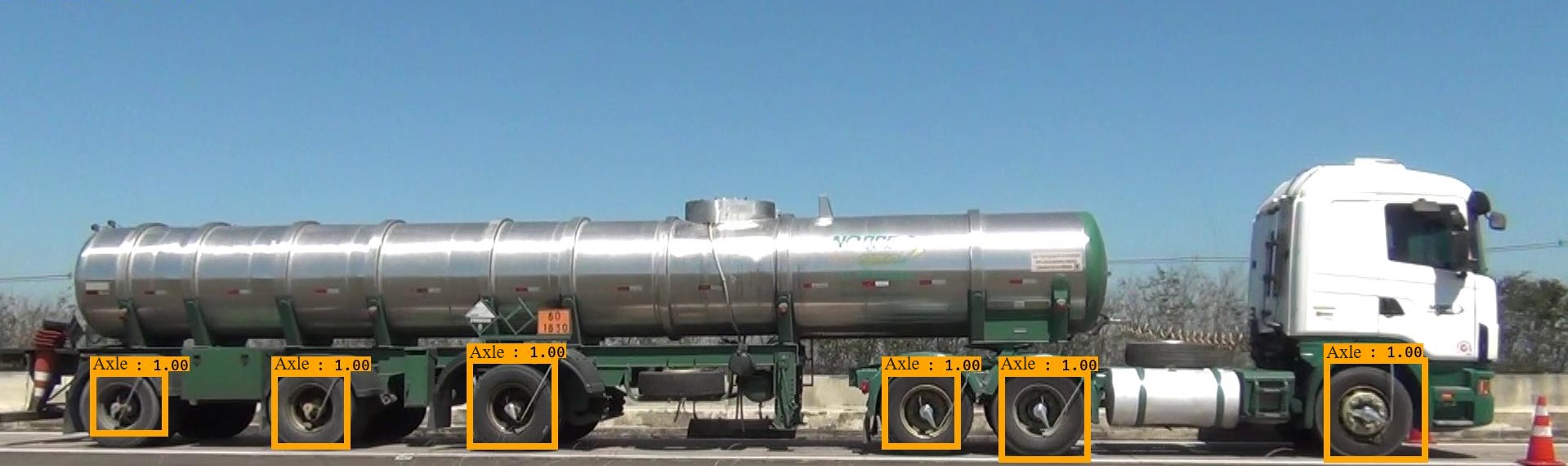}
		\caption{}
		\label{fig:sub:subfigb}
	\end{subfigure}
	\begin{subfigure}[b]{0.985\textwidth}		
		\centering
		\includegraphics[width=\textwidth]{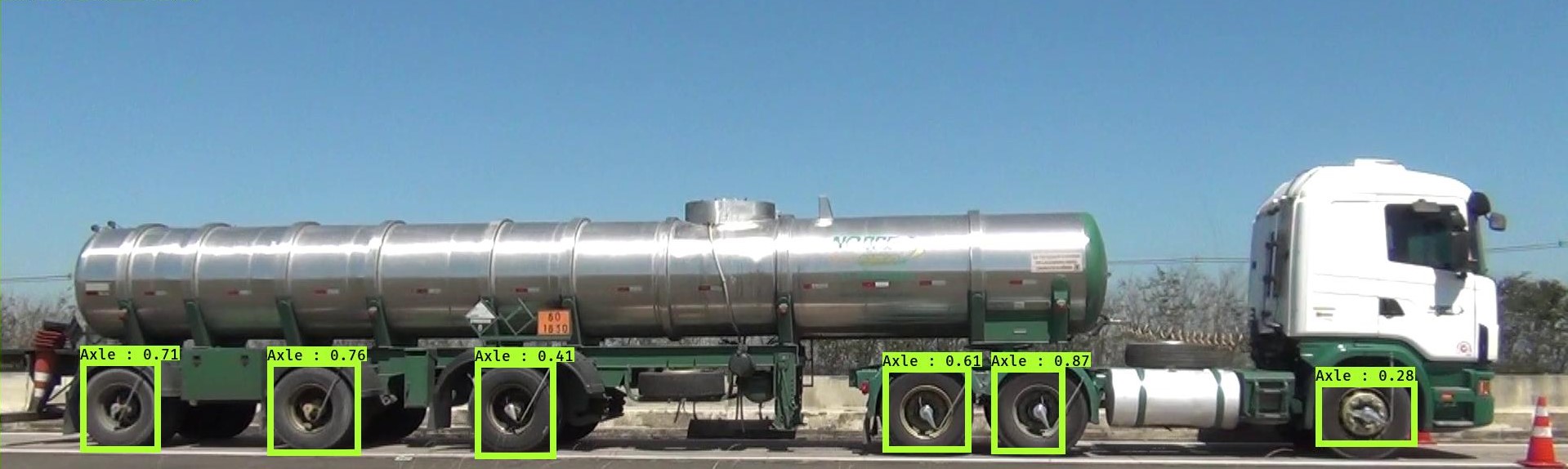}
		\caption{}
	\label{fig:sub:subfigc}
	\end{subfigure}
	\caption{Truck axle detections (a) SSD model (b) Faster R-CNN model (c) YOLOv3-spp model }
	\label{fig:detectionresults}
\end{figure}

Testing was performed with 36 images, all different from the training set, with a total of 119 truck axles. Since all image were from real road traffic, several types of wheel and hubcap configurations were captured. The algorithms were implemented to take advantage of the processing power of the video board, a GeForce GTX 1080, and in Python, with the use of TensorFlow and Keras packages. An example of output detection images is seen on Figure \ref{fig:detectionresults}.

The resulting data was then analyzed by calculating mAP and F1-score for each of the five neural network models since both metrics use detection accuracy and precision. Performance was also calculated based on how many frames per second each model is able to process. 

\section{Results and Discussions}

\begin{figure}[h]
    \centering
	\begin{subfigure}[b]{0.985\textwidth}		
		\centering
		\includegraphics[width=\textwidth]{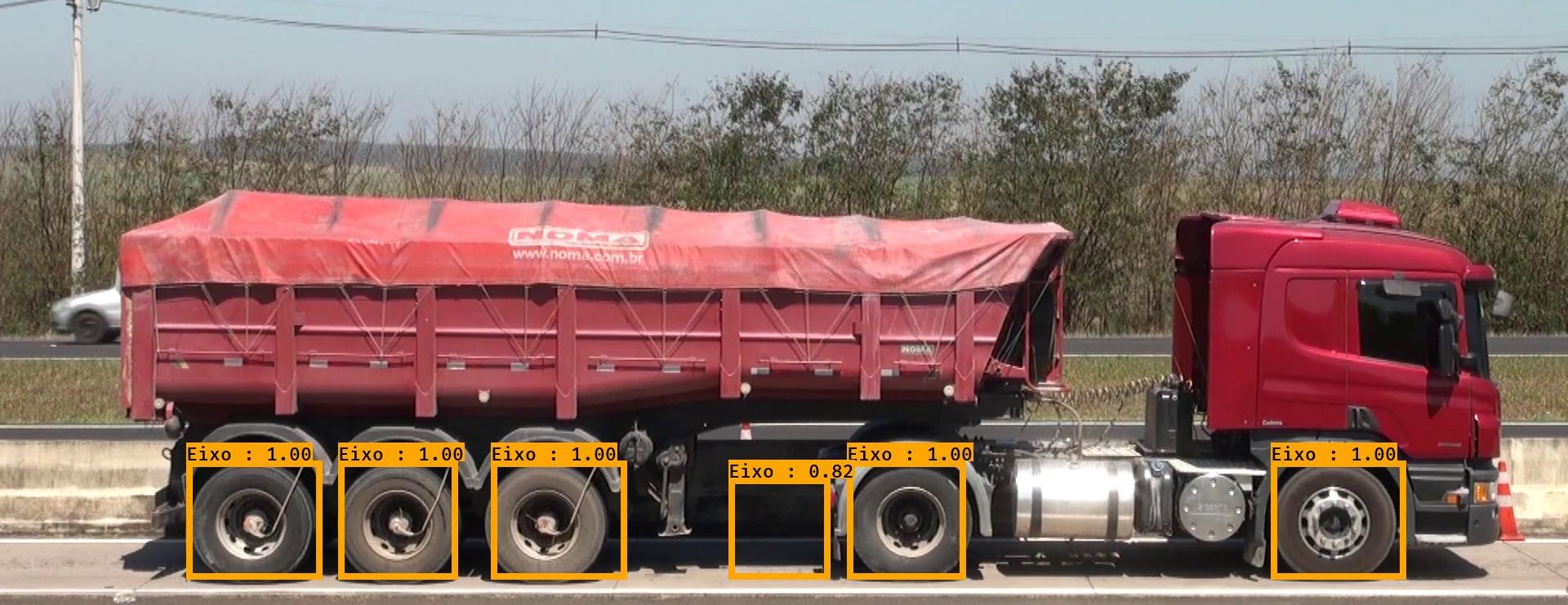}
		\caption{}
		\label{fig:sub:falsopos}
	\end{subfigure}
	\begin{subfigure}[b]{0.985\textwidth}		
		\centering
		\includegraphics[width=\textwidth]{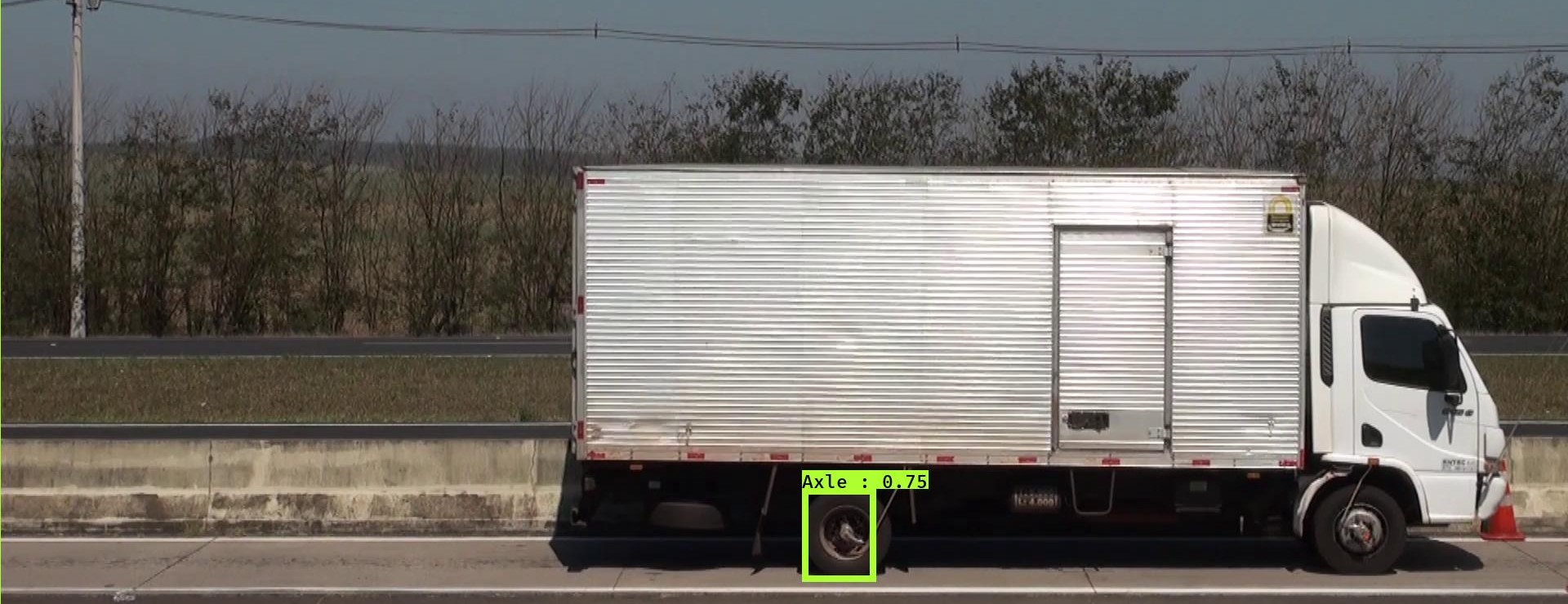}
		\caption{}
		\label{fig:sub:falsoneg}
	\end{subfigure}
	\caption{Detection issues (a) False positive, when an object is detected where there is none (b) False negative when an object is not detected. }
	\label{fig:falseposneg}
\end{figure}

The detection results, which are created after training and applying all models to the testing dataset, can be seen on Table~\ref{table:tab2}. It is possible to notice that all models had similar performances, with high levels of true positives. The difference can be noted on the false positive detections, which happens when a network detects an object where there is none. Faster R-CNN detected 11 false objects, with YOLOv3-tiny close behind, with 9. False negatives, which happen when an object is not found, were also similar between models. It is possible to see examples of false detections on Figure~\ref{fig:falseposneg}. 

\begin{table}[h]
	\centering
	\caption{Detection results for each tested neural network model.}
	\label{table:tab2}
	\begin{tabular}{|c|c|c|c|c|} 
	    \hline
		& \textbf{Resolution} & \textbf{True Positive} & \textbf{False Positive} & \textbf{False Negatives}\\ 
		\hline
		\textbf{YOLOv3-tiny}    & 416x416 & 114 & 9  & 5 \\
		\hline
		\textbf{YOLOv3-416}         & 416x416 & 113 & 3  & 6 \\
		\hline
		\textbf{YOLOv3-spp}     & 608x608 & 117 & 2  & 2 \\
		\hline
		\textbf{FRCNN-resnet50} & 224x224 & 117 & 11 & 2 \\
		\hline
		\textbf{SSD-ssd300}     & 300x300 & 115 & 0  & 4 \\
		\hline
	\end{tabular}
\end{table}

Detection results may appear similar, but considering the fact that the data set is small, with 119 objects, a difference of 11 occurrences is more than 10\%. Although it is not possible to take a decision on which is the best model for this situation solely on these numbers, it is indicative that FRCNN, based on resnet50, and YOLOv3-tiny, may not be the best options for this detection problem. This does not invalidate the algorithms, and parameters can be fine-tuned to improve results, but all tested instances were trained on default values to allow for a better comparison.

By calculating commonly used metrics to analyze object detectors, it is possible to have a deeper look at each model, with metrics that are easier to compare. Table~\ref{table:tab3} has the results of mAP and F1-score, both based on precision and recall of all detections. The difference between these two metrics is that mAP takes into account the localization of the detection, whether the model detected the correct bounding box. The F1-score measures a harmonic mean between precision and recall, i.e., a balanced metric of sensitivity and positive predictive value.

\begin{table}[h]
	\centering
	\caption{Evaluation metrics for each tested neural network model.}
	\label{table:tab3}
	\begin{tabular}{|c|c|c|c|c|} 
	    \hline
		& \textbf{Precision (\%)} & \textbf{Recall (\%)} & \textbf{mAP} & \textbf{F1-score}\\ 
		\hline
		\textbf{YOLOv3-tiny}    & 92.68  & 95.79 & 92.87 & 0.94 \\
		\hline
		\textbf{YOLOv3-416}     & 97.41  & 94.95 & 96.53 & 0.96 \\
		\hline
		\textbf{YOLOv3-spp}     & 98.31  & 98.31 & 98.26 & 0.98 \\
		\hline
		\textbf{FRCNN-resnet50} & 91.40  & 98.31 & 98.32 & 0.95 \\
		\hline
		\textbf{SSD-ssd300}     & 100.00 & 96.63 & 96.64 & 0.98 \\
		\hline
	\end{tabular}
\end{table}

It is possible to notice that FRCNN was more precise than other models when determining the location of the axle (mAP), while YOLOv3-tiny was the worst. This happens because FRCNN uses two steps to identify and locate objects, making it naturally more precise with the localization of objects. Since SSD and YOLOv3 are both based on scanning the image just once to locate objects, it is a trade-off between localization and quickness to detect. YOLOv3-tiny reduces the size of the network in order to speed up the detection and classification process, but in detrimental to object detection, which can be noted in the results. However, both YOLOv3-spp and SSD-ssd300 have similar results to FRCNN, with a difference of just 0.14\% and 1.68\%, respectively. Looking at the F1-score, results are all above 0.94, indicating that all models have excellent performance in correctly detecting objects and not missing many instances.

To analyze the processing speed, it is possible to compare the FPS. Table~\ref{table:tab4} shows how many frames per second each model was capable to process. 

\begin{table}[h]
	\centering
	\caption{FPS for each neural network model.}
	\label{table:tab4}
	\begin{tabular}{|c|c|c|c|c|}
	    \hline
		& \textbf{Model Resolution} & \textbf{FPS}\\ 
		\hline
		\textbf{YOLOv3-tiny}    & 416x416 & 40 \\
		\hline
		\textbf{YOLOv3-416}     & 416x416 & 33 \\
		\hline
		\textbf{YOLOv3-spp}     & 608x608 & 29 \\
		\hline
		\textbf{FRCNN-resnet50} & 224x224 & 21 \\
		\hline
		\textbf{SSD-ssd300}     & 300x300 & 32 \\
		\hline
		\multicolumn{3}{r}{GPU: GeForce GTX 1080.}\\
	\end{tabular}
\end{table}

Despite the fact that processing speed is heavily dependent on available hardware, by comparing in the same environment it is possible to have an idea of which model has the better performance. YOLOv3-tiny, due to it's smaller structure, was capable to process 40 FPS. The worst model was FRCNN, because it is two-step detection weights heavily on performance, with 21 FPS. These results were expected because the specifics of the structure and size of each neural network impact performance. It is important to notice that both SSD and YOLOv3-spp had similar results, processing approximately 30 FPS.

Based on all results, both YOLOv3-spp and SSD-ssd300 are equivalent in detecting truck axles, making them both good candidates for an automatic system. 

\section{Conclusion and Future Works}

We tested three of the most popular architectures of neural networks for object detection, comparing results based both on precision and recall metrics, with mAP and F1-score, and on processing speed, by measuring the FPS. It is possible to notice that YOLOv3 and SSD have similar performances on both metrics and processing speed, allowing for fast and precise object recognition. Both models use a single-step detector, with different strategies to locate the object. Further tests with different weight models are necessary to increase testing variety, and implementing recent versions of each architecture, i.e., YOLOv7 or YOLOv8, is a must for better comparisons.  

Training and testing all networks with the same dataset was important to compare results. The time to train each model was not taken in consideration, even though they were expressive in comparison with testing times. It took approximately 1 week of training for each model, while testing was done in seconds. Training times can be lowered by using better hardware, but with a larger dataset, more processing power will be necessary for smaller processing times.

The dataset size is a limitation of this study. A larger set of images both for training and for testing will improve even further the results. Although they are already good, more examples of axles in different light and weather conditions will improve the robustness of detections, with fewer cases of false negatives and false positives. Since no other truck axle database was found on the internet, sharing the dataset with labels for other research on this topic is a major contribution.

Even with a small dataset for training and testing, detection and classification was achieved with good results, indicating that the use of a specialized neural network, trained for a specific task, is a good alternative to the pre-trained, general-purpose neural networks available on the internet. In addition, not only it is a good alternative, but it can be trained with small datasets. Since acquiring enough data may be a deterrent for researchers, another major contribution of this paper is to indicate that using small datasets to train specialized neural networks is possible, and with good results.

For the objective of detecting truck axles, an object that is neglected in pre-trained neural networks but is important in several use cases for planning and operating transport systems, this paper was able to investigate and compare different architectures, with the production of a publicly available dataset \cite{leandro_arab_marcomini_2021_5744737}, which will be improved with more images, and the implementation of all tested neural networks, available in \cite{axle}.

\section{Acknowledgments}

This study was financed in part by the Coordenação de Aperfeiçoamento de Pessoal de Nível Superior - Brasil (CAPES) - Finance Code 001. This research also received financial support from the National Council for Scientific and Technological Development (CNPq), project no. 436954/2018-4.

\bibliographystyle{ieeetr}
\bibliography{ref}

\end{document}